\documentclass[lettersize,journal]{IEEEtran}
\usepackage{amsmath,amsfonts}
\usepackage{algorithmic}
\usepackage{algorithm}
\usepackage{array}
\usepackage[caption=false,font=normalsize,labelfont=sf,textfont=sf]{subfig}
\usepackage{textcomp}
\usepackage{stfloats}
\usepackage{url}
\usepackage{verbatim}
\usepackage{graphicx}
\usepackage{cite}
\usepackage{soul}
\usepackage{color}
\usepackage{multirow}
\usepackage{utfsym}
\usepackage{graphicx}
\usepackage{amsmath}
\usepackage{amssymb}
\usepackage{booktabs}
\usepackage{colortbl}
\usepackage{fontawesome}
\usepackage{mathtools}
\usepackage{arydshln} 
\usepackage{hyperref}
\usepackage{threeparttable}
\newcommand{\W}[1]{{\scriptsize\itshape\textsc{#1}}}
\begin{document}

\title{Contextualized Representation Learning for Effective \\ Human-Object Interaction Detection}

\author{Zhehao Li, Yucheng Qian, Chong Wang \textsuperscript{\faEnvelope}, Yinghao Lu, Zhihao Yang and Jiafei Wu
\thanks{Manuscript received 3 October 2025; This work was supported by the Ningbo Municipal Natural Science Foundation of China (No. 2022J114), National Natural Science Foundation of China (No. 62271274), Ningbo S\&T Project (No.2024Z004) and Ningbo Major Research and Development Plan Project (No.2023Z225).}
\thanks{Zhehao Li, Chong Wang, Yinghao Lu, and Zhihao Yang are with the Faculty of Electrical Engineering and Computer Science, Ningbo University, Ningbo, Zhejiang 315211, China, E-mail: lllzzzhhh1019@163.com, wangchong@nbu.edu.cn.}
\thanks{Yucheng Qian is with Nanjing University. Email: qianycqq@163.com.}
\thanks{Jiafei Wu is with Zhejiang Lab, Hangzhou, China. E-mail: {wujiafei@zhejianglab.com}.}
\thanks{\faEnvelope \ Corresponding Author: Chong Wang.}}

\markboth{Journal of \LaTeX\ Class Files,~Vol.~14, No.~8, August~2021}%
{Shell \MakeLowercase{\textit{et al.}}: A Sample Article Using IEEEtran.cls for IEEE Journals}

\IEEEpubid{0000--0000/00\$00.00~\copyright~2021 IEEE}

\maketitle

\begin{abstract}
Human-Object Interaction (HOI) detection aims to simultaneously localize human-object pairs and recognize their interactions. While recent two-stage approaches have made significant progress, they still face challenges due to incomplete context modeling. In this work, we introduce a Contextualized Representation Learning that integrates both affordance-guided reasoning and contextual prompts with visual cues to better capture complex interactions. We enhance the conventional HOI detection framework by expanding it beyond simple human-object pairs to include multivariate relationships involving auxiliary entities like tools. Specifically, we explicitly model the functional role (affordance) of these auxiliary objects through triplet structures $<$human, tool, object$>$. This enables our model to identify tool-dependent interactions such as “filling”. Furthermore, the learnable prompt is enriched with instance categories and subsequently integrated with contextual visual features using an attention mechanism. This process aligns language with image content at both global and regional levels. These contextualized representations equip the model with enriched relational cues for more reliable reasoning over complex, context-dependent interactions. Our proposed method demonstrates superior performance on both the HICO-Det and V-COCO datasets in most scenarios. The source code is available at \url{https://github.com/lzzhhh1019/CRL}.
\end{abstract}

\begin{IEEEkeywords}
Human-Object Interaction Detection, Two-Stage Network, Prompt Learning, Attention Mechanism.
\end{IEEEkeywords}

\section{Introduction}
\label{sec:intro}

\IEEEPARstart{H}{uman-Object} Interaction (HOI) detection is a fundamental and challenging task in visual recognition that aims to model complex semantic relationships between humans and surrounding objects in a visual scene, ultimately recognizing triplets of the form $<$\textit{human, action, object}$>$. It requires a simultaneous understanding of visual features, spatial configurations and contextual semantics to effectively identify all valid interaction patterns within given images. Therefore, other high-level semantic understanding tasks, such as activity recognition \cite{liu2022attention,xu2023towards,action_recognition_2025_CVPR} and video comprehension \cite{tao2024feature,wang2024gist,shi2024commonsense} can benefit from HOI.

\begin{figure}[htbp]
\centering
\includegraphics[width=\columnwidth]{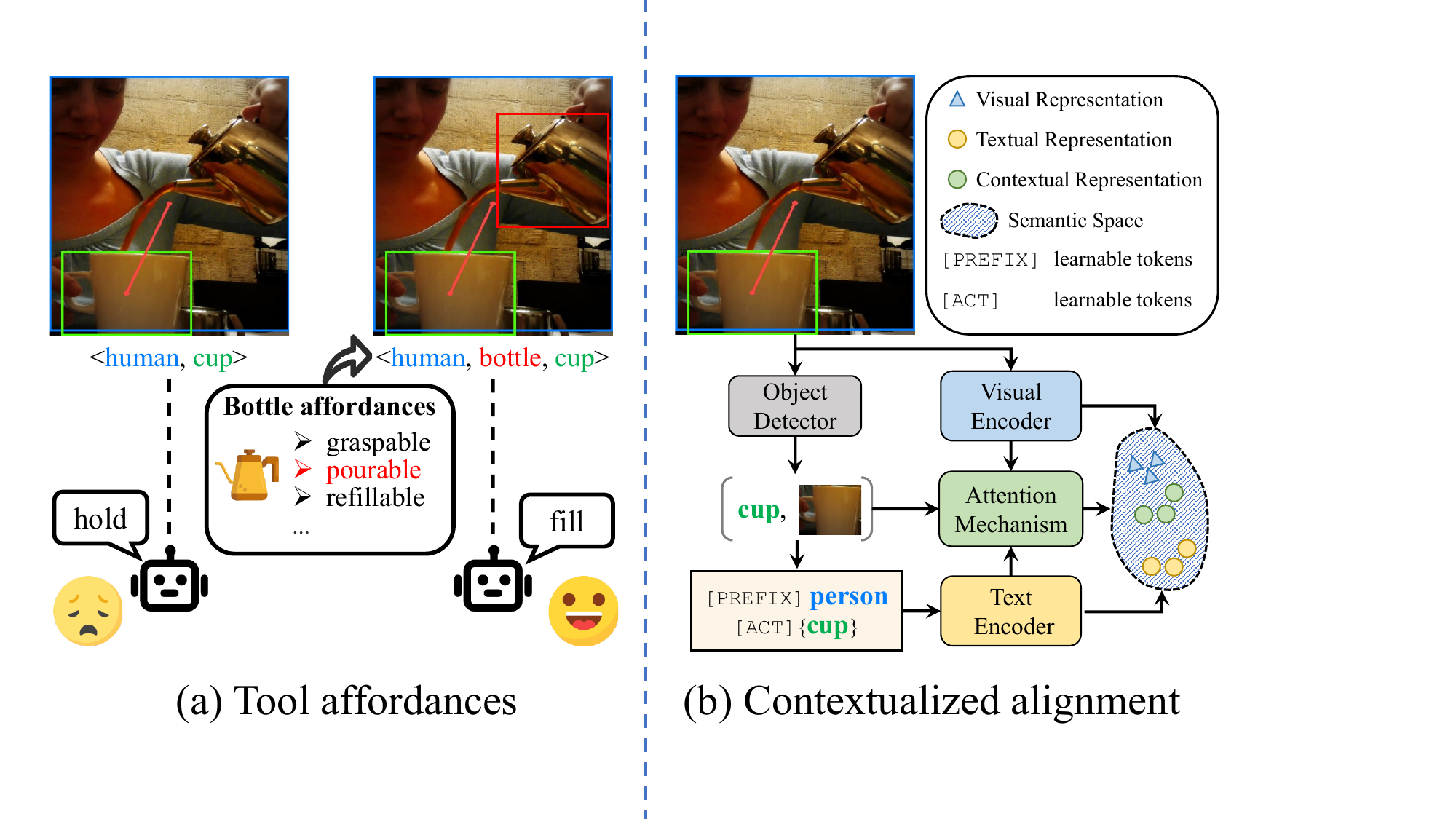}
\caption{Contextualized representations in HOI. (a) Tool affordances (e.g., bottle's pourable) help distinguish complex interactions (e.g., human \textbf{fill} cup) from direct human–object relations (e.g., human hold cup). (b) Contextualized alignment with instance categories (e.g., cup) and corresponding visual features narrows down the potential actions to relevant ones (e.g., fill, hold).}
\label{fig:fig1}
\end{figure}

Existing HOI detection approaches can generally be divided into one-stage and two-stage methods. One-stage methods perform object detection and interaction prediction simultaneously, but adapting their decoders for various HOIs often incurs substantial training costs, sometimes requiring hundreds of GPU hours. On the other hand, two-stage methods leverage pre-trained object detectors to identify humans and objects before explicitly establishing relationships between them for interaction prediction. Thanks to its flexibility, the two-stage approach has quickly become the preferred paradigm.
However, traditional two-stage HOI approaches have primarily relied on features derived from human-object pairs, such as human pose \cite{wu2024exploring} or spatial information \cite{zhang2022efficient}. In real-world interactions, many scenarios are not solely defined by these pairs. They may also involve auxiliary tools or objects that play a crucial role in the interaction. While prior HOI studies \cite{hou2021affordance} have leveraged object affordances, such modeling usually remains tied to the acted-on object and does not explicitly account for tool affordances. In contrast, we model the functional role of tools within the $<$human, tool, object$>$ ternary. For example, as illustrated in Fig. \ref{fig:fig1}(a), when a person fills a cup using a bottle, understanding the bottle's affordance (i.e., its potential function like being pourable) offers significant insights into deciphering the nature of the interaction.

\IEEEpubidadjcol

Meanwhile, recent advancements in pre-trained language models like BERT \cite{devlin2019bert} and GPT \cite{brown2020language} offer rich semantic insights for understanding HOIs. Prompt Learning, as an efficient method for finetuning language models, has been extened to visual tasks by CoOp \cite{zhou2022learning}. Many studies have as integrated Prompt Learning into the HOI task \cite{wang2022learning,lei2024exploring,lei2024ez}. However, these approaches rely exclusively on textual data and do not integrate image information from specific samples. As a result, crucial contextual visual cues are left untapped, which are essential for accurately capturing interactions.

It's worth noting that both the affordances and the text-related appearance of a specific object act as vital forms of contextualized representation, essential for identifying complex interactions. Despite the importance, their applications in HOI has not been thoroughly explored. This gap in research greatly motivates us to delve deeper into this area. In this work, we propose a new two-stage HOI detection framework, named Contextualized Representation Learning (CRL). To be specific, unlike traditional human-object pairs, multivariate relationships (unary, binary and ternary ones) are modeled to capture tool-mediated interactions through affordance-guided triplets $<$human, \textbf{tool}, object$>$. In parallel, the contextualized representations of detected instances are injected into the process of prompt learning. Specifically, as illustrated in Fig. \ref{fig:fig1}(b), we extract localized visual features from the corresponding instance regions and refine the prompt representation through a cross-attention mechanism with VLM’s global image embedding. This design allows the prompt to be conditioned on both textual semantics and instance-level visual cues, improving its capacity to capture context-dependent interactions.

To sum up, our contributions are two-fold:
\begin{itemize}
\item We are the first to emphasize the role of tools in improving HOI understanding. Guided by the concept of affordances, we propose a multivariate relationship modeling framework that introduces $<$human, \textbf{tool}, object$>$ triplets to capture complex tool-mediated interactions beyond conventional human-object pairs.
\item We introduce a contextualized prompt learning module that incorporates detected object categories and their corresponding visual features into the prompt, enabling contextual alignment between textual and visual modalities at both semantic and instance levels.
\end{itemize}

\section{Related Work}
\subsection{HOI Detection}

Existing Human-Object Interaction (HOI) detection approaches can be divided into one-stage and two-stage methods. One-stage methods \cite{zou2021end,kim2021hotr,tamura2021qpic,xie2023category,fang2023hodn} simultaneously predict human and object boxes, categories, and interaction classes. Recent advancements in one-stage detectors, particularly those utilizing transformer architectures \cite{zhang2021mining,kim2023relational,zhang2024plug,gao2024dual}, have shown promising performance. This end-to-end approach simplifies the inference process of HOI. However, it imposes heavy computational costs during training, which can limit its practicality. 

To alleviate this issue, two-stage methods \cite{he2021exploiting,liu2022interactiveness,wu2022mining,zhang2022exploring,zheng2023open,guo2024unseen} undergone significant development recently. 
They typically rely on pre-trained detectors (e.g., DETR) to generate object proposals, focusing more on extracting interaction context from candidate human-object pairs. For example, UPT \cite{zhang2022efficient} first detects all humans and objects (i.e., unary features) and then applies self-attention to unary features and human-object pairs, effectively enhancing the confidence of positive samples. PViC \cite{zhang2023exploring} incorporates image features back into the representation of human-object pairs via cross-attention to compensate for missing contextual information. Nonetheless, traditional methods often treat all objects equally, failing to account for the role that tools play in interactions. In this work, we explore the concept of tool affordances to more effectively capture these relational cues.

\subsection{Vision-Language Models in HOI}
To obtain more effective HOI representations, several studies \cite{liao2022gen} explore knowledge transfer from vision-language pre-trained models (e.g., CLIP \cite{radford2021learning}). This approach not only enriches the learned representations but also improves the model's capability in HOI recognition. GEN-VLKT \cite{liao2022gen} utilizes CLIP knowledge for interaction classification and the distillation of visual features. HOICLIP \cite{ning2023hoiclip} uses the features obtained by the VLM visual encoder and proposes a new transfer strategy that uses visual semantic algorithms to represent action. ViPLO \cite{zou2021end} adopts the Vision Transformer (ViT) from CLIP as its backbone and introduces a pose-conditioned graph to capture the local features of human joints. ADA-CM \cite{lei2023efficient} develops a concept-guided memory mechanism to represent visual embeddings and semantic knowledge simultaneously. The success of VLMs opens up new avenues for HOI detection. CLIP4HOI\cite{mao2023clip4hoi} utilizes CLIP's vision-language knowledge to enhance human-object interaction detection by decoupling human and object detection and adapting CLIP into a fine-grained classifier for better interaction discrimination.

\subsection{Prompt Learning in HOI}
Prompt learning has become very popular for fine-tuning VLMs on downstream tasks. Context Optimization (CoOp) \cite{zhou2022learning} encodes prompt context as learnable vectors, achieving strong performance with only a few labeled samples. To better adapt to downstream tasks, Conditional Context Optimization (CoCoOp) \cite{zhou2022conditional} builds upon CoOp by employing a lightweight network to generate input-conditional context tokens for each image. THID \cite{wang2022learning} is the first to propose the use of learnable language prompts for the HOI detection task. CMMP \cite{lei2024exploring} introduces decoupled multi-modal prompts for spatial-aware HOI detection, separating visual feature extraction and interaction classification to reduce error propagation. EZ-HOI \cite{lei2024ez} leverages LLM-generated class descriptions to guide prompt learning. 

However, most existing prompt learning methods for HOI tasks neglect instance-specific visual details and fail to adapt contextually. To overcome these limitations, we integrate contextualized visual representations to create entity-aware prompts, thereby enhancing the model's understanding of visual content.

\begin{figure*}[htbp]
\centering
\includegraphics[width=\textwidth]{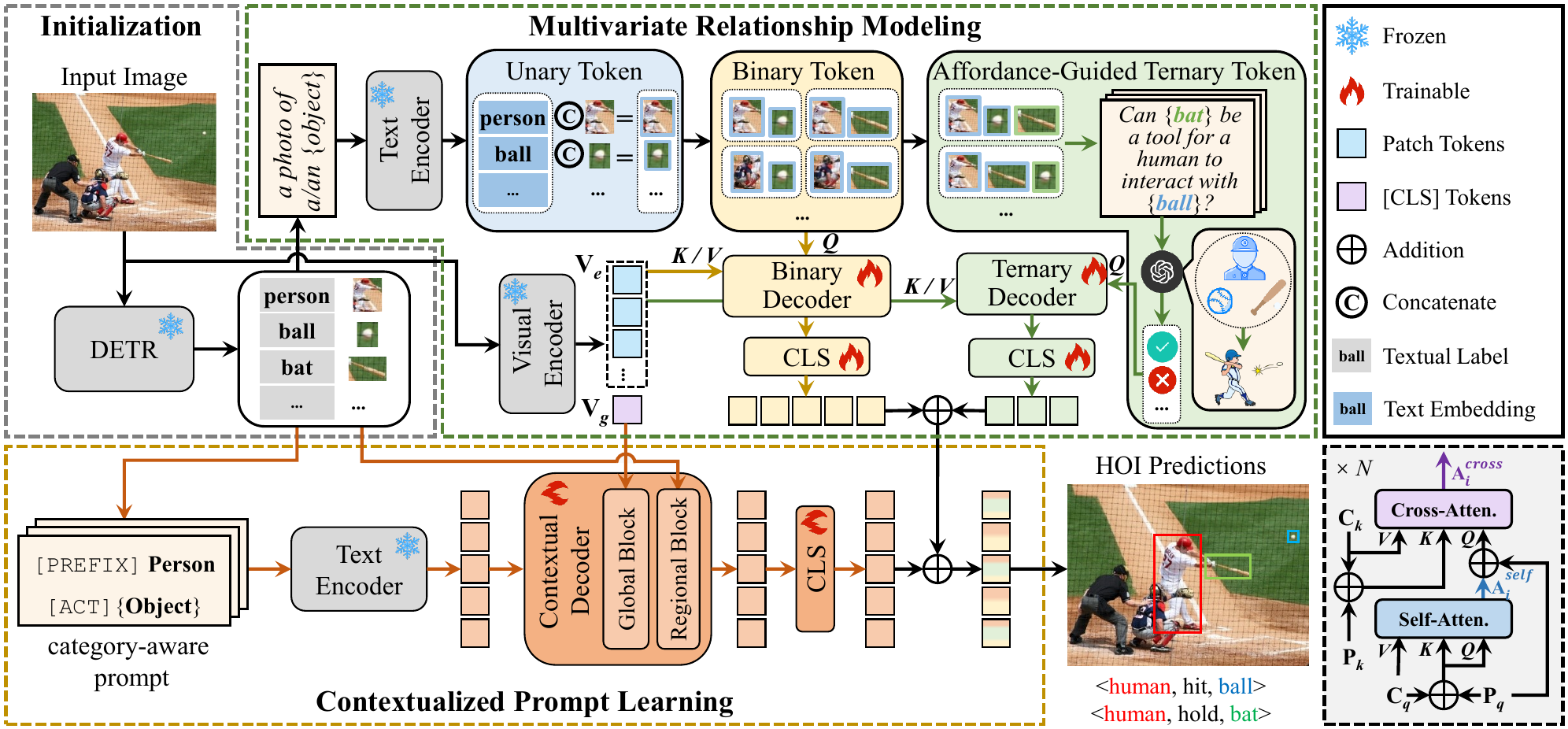}
\caption{Overall architecture of our Contextualized Representation Learning Network, consisting of Multivariate Relationship Modeling (MRM) and Contextualized Prompt Learning (CPL). MRM constructs unary, binary and ternary token sets from regional features to model HOIs. CPL builds a category-aware learnable prompt, fused with diverse contextual visual features. Their combined outputs are utilized for interaction prediction. The structure of the binary/ternary/contextual decoder is shown in the bottom right. } 
\label{fig:fig2}
\end{figure*}

\section{Method}

\subsection{Preliminary}

Our proposed model builds upon the traditional two-stage methodology for Human-Object Interaction (HOI) detection, which first focuses on detecting instances and subsequently classifies the interactions between them. In the first stage, as shown in the upper-left part of Fig. \ref{fig:fig2}, we employ an off-the-shelf object detector, such as DETR \cite{carion2020end}, 
to extract all instances, including humans and objects. As a result, a set of instances $\mathcal{Z} = \{z_i\}_{i=1}^n$ can be constructed with each detection $z_i \, = \, (\mathbf{b}_i, s_i, c_i, \mathbf{u}_i)$, consisting of the box coordinates $\mathbf{b}_i \in \mathbb{R}^4$, the confidence score $s_i \in [0, 1]$, the detected instance $c_i \in \mathcal{O}$ for the category set $\mathcal{O}$, and the unary instance feature $\mathbf{u}_i \in \mathbb{R}^C$. $C$ is the embedding dimension, $n$ is the number of detected instances. 

In the second stage, the visual and semantic features of the detected instance set $\mathcal{Z}$ are utilized to exploit the interaction information. In this work, as shown in Fig. \ref{fig:fig2}, we focus on modeling relationships at three distinct levels: unary (involving individual entities), binary (encompassing human-object interactions) and ternary (covering human-tool-object dynamics) ones. This enables affordance-guided interaction modeling, which allows the model to effectively capture the progressively complex patterns inherent in HOIs. Moreover, we further extract the contextualized representation from $\mathcal{Z}$ to improve prompt learning. Specifically, a category-aware prompt is fused with contextual unary features and CLIP’s global visual features via an attention mechanism.

\subsection{Multivariate Relationship Modeling}

\textbf{Unary Association.} To establish the multivariate relationships among the detected instances in $\mathcal{Z}$, we start with the most straightforward scenario, i.e., examining the visual and textual representation of each instance (whether human or object) individually. In this unary association, its visual feature is denoted as a unary token $\mathbf{u}_i$, forming the set $\mathcal{U} = \{\mathbf{u}_i\}_{i=1}^n$. Meanwhile, the corresponding semantic feature $\mathbf{e}_i$ for each instance is obtained through the text encoder of a vision-language model (VLM). This process employs the prompt template “\textit{a photo of a/an \{object\}}”, where \{object\} is the instance category $c_i$ detected in the first stage. Then, such category-level semantic information is combined with visual features to obtain an enhanced unary token $\mathbf{u}_i'$ as,
\begin{equation}
\mathbf{u}_{i}'\ = \operatorname{MLP}\big( \operatorname{Concat}(\mathbf{u}_i,\mathbf{e}_{i}) \big),
\end{equation}
where $\operatorname{MLP}(*)$ is a multi-layer perceptron used to project the concatenated features into the target embedding space.

{\textbf{Binary Connection.}} Based on the unary association, we continue to investigate the binary connection between different instances. Inspired by UPT \cite{zhang2022efficient}, a binary token set $\mathcal{G}$ = $\{\mathbf{g}_l\}_{l=1}^m$ can be constructed from the unary one $\mathcal{U}$ as follows,
\begin{equation}
\begin{aligned}
    \mathbf{g}_l = & \operatorname{MLP}\big( 
        \operatorname{Concat}(\mathbf{u}_i', \mathbf{u}_j') 
        \big), \\
    &\text{s.t. } \left\{
    \begin{array}{l}
    i \neq j, \\
    c_i = \text{``human''}.
    \end{array}
    \right.
\end{aligned}
\label{eq:ho_pairwise}
\end{equation}
As the fusion of human token $\mathbf{u}_{i}'$ and object token $\mathbf{u}_{j}'$, $\mathbf{g}_l \in \mathbb{R}^{D}$ is fed into a binary decoder for interaction classification.

Following the setup in DETR \cite{carion2020end}, the structure of our binary decoder is presented in the right lower corner of Fig. \ref{fig:fig2}. It includes $N$ blocks, each with cascaded self- and cross-attention layers, designed to process four input, namely content query $\mathbf{C}_q$, positional query $\mathbf{P}_q$, content key $\mathbf{C}_k$, and positional key $\mathbf{P}_k$. The attention layers follow the standard process $\operatorname{Atten}(*)$ as,
\begin{equation}
\operatorname{Atten(\mathbf{Q},\mathbf{K},\mathbf{V}} ) = \operatorname{Norm}\left(
\sigma\left(\frac{\mathbf{Q} \mathbf{K}^\top }{\sqrt{d_k}}\right) \mathbf{V} + \mathbf{V}
\right),
\end{equation}
where $\mathbf{Q}$, $\mathbf{K}$, $\mathbf{V}$ are the query, key and value tokens. $\operatorname{Norm}(*)$ and $\sigma(*)$ denotes the layer normalization and softmax function, respectively.




All tokens $\mathbf{g}_l \in \mathbb{R}^{D} $ are concatenated into a matrix $\mathbf{G}_0 \in \mathbb{R}^{m \times D}$, which is used as the content query $\mathbf{C}_q$. 
The corresponding positional matrix $\mathbf{X} \in \mathbb{R}^{m \times D}$ is used as the positional query $\mathbf{P}_q$.
It is extracted from all bounding box pair $(\mathbf{b}_i, \mathbf{b}_j)$, by combining various attributes.
$D$ is the embedding dimension and $m$ is the number of human-object pairs. Then, the output of self-attention process $\mathbf{A}_i^{\textit{b-self}}$ in the $i$-th binary decoder block can be expressed as follows,
\begin{equation}
    \mathbf{A}_i^{\textit{b-self}} =
    \operatorname{Atten} \left(
    \mathbf{G}_{i-1}+\mathbf{X},\mathbf{G}_{i-1}+\mathbf{X},\mathbf{G}_{i-1}
    \right),    
\end{equation}   
where $i=1,2,...,N$,  
$\mathbf{G}_{i-1}$ is the outputs of the previous decoder block.

To enhance interaction modeling, we further employ the visual encoder \textit{VisEnc}$(*)$ of VLMs to extract spatial features $\mathbf{V}_e \in \mathbb{R}^{H' \times W' \times D}$ from the whole image, accompanying by the position embedding $\mathbf{S} \in \mathbb{R}^{H' \times W' \times D}$. 
Here, $\mathbf{V}_e$ serves as content keys and values, while $\mathbf{S}$ acts as positional keys. Thus, the output of cross-attention layer $\mathbf{A}_i^{\textit{b-cross}}$ in the $i$-th block can be formulated as,
\begin{equation}
    \mathbf{A}_i^{\textit{b-cross}} = \operatorname{Atten} \left(
    \mathbf{A}_i^{\textit{b-self}}+\mathbf{X},\mathbf{V}_e + \mathbf{S},\mathbf{V}_e
    \right).
\end{equation}
The final output $\mathbf{G}_i$ of $i$-th block is obtained as,
\begin{equation}
    \mathbf{G}_i = \operatorname{Norm} \left(\mathbf{A}_i^{\textit{b-cross}}+\operatorname{FFN}
    \left(\mathbf{A}_i^{\textit{b-cross}}\right)
    \right),
\end{equation}
where $\operatorname{FFN}(*)$ denotes a feed-forward network. 
The output of the last block $\mathbf{G}_N$ is then fed into a linear layer to predict the classification logits $\tilde{\mathbf{y}} \in \mathbb{R}^{m \times c}$ for human-object pairs’ interactions, where 
$c$ is the number of action categories. 


{\textbf{Ternary Relationship.}} Beyond the binary connections, we further explore the ternary relationships within $\mathcal{Z}$ by incorporating the concept of object affordances into the HOI task. To determine which objects can serve as functional tools in HOIs, we query a large language model (LLM) using a prompt of “\textit{Can \{\texttt{X}\} be a tool for a human to interact with \{\texttt{Y}\}?}”, where \texttt{X} and \texttt{Y} represent different object categories (e.g., bat and ball). Although the prompt does not explicitly mention affordances, the LLM exhibits a strong implicit understanding of object functionality and affordance-related reasoning. Based on LLM’s responses, we construct a set of object-tool pairs, which are stored in the knowledge bank $\mathcal{B}$ as ordered pairs of the form $<$\texttt{Y}, \texttt{X}$>$. To improve efficiency, these identified object-tool pairs are curated offline and used as external knowledge during training and inference.




Similar to the binary token set, a ternary token set $\mathcal{T}$ = $\{\mathbf{t}_o\}_{o=1}^r$ of size $r$ is constructed as follows,
\begin{equation}
\begin{aligned}
    \mathbf{t}_o &= \operatorname{MLP}\big( \operatorname{Concat}(\mathbf{u}_i', \mathbf{u}_j', \mathbf{u}_k') \big), \\
    &\text{s.t. } \left\{
    \begin{array}{l}
    i \neq j,\, j \neq k,\, i \neq k, \\
    c_i = \text{``human''}, \\
    <c_j, c_k> \in \mathcal{B}.
    \end{array}
    \right.
\end{aligned}
\label{eq:ternary}
\end{equation} 


A ternary decoder, designed with a structure similar  to the binary decoder but having different coefficients, is developed to leverage affordance-guided interaction from $\mathcal{T}$.
Noting that each ternary token $ \mathbf{t}_o \in \mathbb{R}^{D} $ is associated with a triplet consisting of the human token $\mathbf{u}_i'$, object token $\mathbf{u}_j'$ and tool token $ \mathbf{u}_k' $. Based on their corresponding bounding box triplet $(\mathbf{b}_i, \mathbf{b}_j, \mathbf{b}_k)$, we compute the relationships between each pair of tokens, i.e., human-object, human-tool, and object-tool. These pairwise features are then concatenated and passed through an MLP to form 
the ternary positional matrix $ \mathbf{W} \in \mathbb{R}^{r \times D} $, 
which serves as the positional query $\mathbf{P}_q$. Then, the ternary matrix $ \mathbf{T}_{0} \in \mathbb{R}^{r \times D} $, concatenated by all $ \mathbf{t}_o $, is used as the content query $\mathbf{C}_q$. The output of self-attention $\mathbf{A}_i^{\textit{t-self}}$ in $i$-th ternary decoder block is,
\begin{equation}
    \mathbf{A}_i^{\textit{t-self}} = \operatorname{Atten} \left(
    \mathbf{T}_{i-1}+\mathbf{W},\mathbf{T}_{i-1}+\mathbf{W},\mathbf{T}_{i-1}
    \right),
\end{equation} 
where $i=1,2,...,N$, and $\mathbf{T}_{i-1}$ represents the output from the previous decoder block.

We continue to use the VLM's spatial features \( \mathbf{V}_e \) and position embeddings \( \mathbf{S} \) as content keys and positional keys. Thus, ternary cross-attention $\mathbf{A}_i^{\textit{t-cross}}$ in $i$-th block is,
\begin{equation}
    \mathbf{A}_i^{\textit{t-cross}} = \operatorname{Atten} \left(
    \mathbf{A}_i^{\textit{t-self}}+\mathbf{W},\mathbf{V}_e + \mathbf{S},\mathbf{V}_e
    \right).
\end{equation} 
The final output $\mathbf{T}_{i}$ of $\textit{i}$-th ternary decoder block is given as,
\begin{equation}
    \mathbf{T}_i = \operatorname{Norm} \left(\mathbf{A}_i^{\textit{t-cross}}+\operatorname{FFN}
    \left(\mathbf{A}_i^{\textit{t-cross}}\right)
    \right).
\end{equation}
Similarly, $\mathbf{T}_N$ from the final block of ternary decoder is then passed through a classifier to generate the affordance-guided prediction logits $\mathbf{y}' \in \mathbb{R}^{r \times c} $.



Typically, the number of human-object pairs, denoted as $m$, does not match the number of human-tool-object triplets, represented by $r$.
As a result, the logits $\tilde{\mathbf{y}}$ and $\mathbf{y}'$ might have different dimensions. It is necessary to effectively fuse them together, ensuring that the disparate information from binary and ternary sets can be integrated in a meaningful way. 
For the $l$-th human-object pair, whenever the triplet for $\mathbf{t}_o$ includes this specific pair, the ternary logits $\mathbf{y}'_o$ shall contribute to classifying that interaction.  Considering that different tools might be involved in such interactions, we define the subset of ternary tokens associated with the same $l$-th human-object pair as $\mathcal{K}_l$.
Thus, the refined interaction logits $\hat{\mathbf{y}}$ are defined as follows,
\begin{equation}
    \hat{\mathbf{y}}_{l} = \tilde{\mathbf{y}}_{l} + \alpha \cdot \sum_{\mathbf{t}_o \in \mathcal{K}_l} \mathbf{y}'_o,
\end{equation}
where $\alpha$ is the weighting parameter. 

\subsection{Contextualized Prompt Learning}
To fulfill the missing piece of corresponding visual information in conventional prompt learning, we propose Contextualized Prompt Learning (CPL) to incorporate specific regional features $\mathbf{D} \in \mathbb{R}^{m \times C'}$. As shown in the lower part of Fig. \ref{fig:fig2}, these contextual representations are integrated with the global visual context $\mathbf{V}_g$ obtained by projecting $\mathbf{V}_e$ through a projection layer,
in order to enhance the effect of learnable text tokens for better interaction prediction.




The contextual features $ \mathbf{D} \in \mathbb{R}^{m \times C'} $ is the concatenation of all $\mathbf{d}_l \in \mathbb{R}^{C'}$, using 
pure visual features of candidate human-object pairs as follows,
\begin{equation}
\begin{aligned}
    \mathbf{d}_l = & \operatorname{MLP}\big( 
        \operatorname{Concat}(\mathbf{u}_i, \mathbf{u}_j) 
        \big). \\
\end{aligned}
\end{equation}
The corresponding object labels (i.e., $c_j$) are used to construct a prompt in the form of “\textit{\texttt{[PREFIX]} person \texttt{[ACT]} \{object\}}”, where \texttt{[PREFIX]} and \texttt{[ACT]} can be 
a sequence of learnable tokens, i.e., $[\mathbf{v}]_1$$[\mathbf{v}]_2$...$[\mathbf{v}]_A$. Each $[\mathbf{v}]_a$ ($a \in \{1,..., A \}$) is a vector with the same dimension as word embeddings (i.e., 512 or 768 for CLIP), and $A$ is a hyperparameter specifying the number of learnable tokens. Then, they are passed through the VLM's text encoder and a projection layer to produce the textual features $\mathbf{M}_{0} \in \mathbb{R}^{m \times C'}$. 

A new contextual decoder, whose block shares the same structure as the binary and ternary ones, is proposed to fuse regional and global contextual visual information (i.e., $\mathbf{D}$ and $\mathbf{V}_g$) with adaptive textual features $\mathbf{M}_{0}$ for interaction reasoning. 
Specifically, two decoder blocks (global and regional), i.e., $N$ = 2, are designed to sequentially inject $\mathbf{V}_g$ and $\mathbf{D}$ at the cross-attention layer. 
Meanwhile, the positional queries or keys are set to $\mathbf{0}$, since the location of instances is less relevant to the prompt. 
Given $\mathbf{M}_{0}$ as the input, the global visual context $\mathbf{V}_g$ is utilized to form the context key and value at the first cross-attention layer. It is repeated $m$ times to obtain $\mathbf{V}'_g \in \mathbb{R}^{m \times C'}$, aligned with the input's dimensions. Thus, the output $\mathbf{M}_1$ of the first block  can be formulated as,
\begin{equation}
    \mathbf{A}_1^{\textit{u-cross}} = \operatorname{Atten} \left(
    \operatorname{Atten} \left( \mathbf{M}_{0}
    ,\mathbf{M}_{0},\mathbf{M}_{0}\right),
    \mathbf{V}'_g,\mathbf{V}'_g
    \right),
\end{equation} 
\begin{equation}
    \mathbf{M}_1 = \operatorname{Norm} \left(\mathbf{A}_1^{\textit{u-cross}}+\operatorname{FFN}
    \left(\mathbf{A}_1^{\textit{u-cross}}\right)
    \right).
\end{equation} 
The regional features $\mathbf{D}$ are applied to further refine the contextualized representation at the second block as, 
\begin{equation}
    \mathbf{A}_2^{\textit{u-cross}} = \operatorname{Atten} \left(
    \operatorname{Atten} \left( \mathbf{M}_{1}
    ,\mathbf{M}_{1},\mathbf{M}_{1}\right)
    ,\mathbf{D},\mathbf{D}
    \right) .
\end{equation}
The final output $\mathbf{M}_{2}$ of the contextual decoder is given as,
\begin{equation}
    \mathbf{M}_2 = \operatorname{Norm} \left(\mathbf{A}_2^{\textit{u-cross}}+\operatorname{FFN}
    \left(\mathbf{A}_2^{\textit{u-cross}}\right)
    \right).
\end{equation}
By integrating the overall scene context and fine-grained instance details, feature representation becomes more comprehensive than methods that rely solely on text-based prompts, which helps to understand complex interactions.

Finally, we train a classifier for $\mathbf{M}_2$ to output the semantic interaction logits $\dot{\mathbf{y}} \in \mathbb{R}^{m \times c}$. It is then integrated with the previously refined interaction logits $\hat{\mathbf{y}}$ as,
\begin{equation}
    \hat{\mathbf{y}}'_l = \hat{\mathbf{y}}_l + \beta \cdot \dot{\mathbf{y}}_l ,
\end{equation}
where $\beta$ is the weighting parameter. 

\subsection{Training and Inference}
\label{subsec:trian_inference}
\textbf{Training.} During training, the Focal Loss (\textbf{FL}) is used on the predicted action logits
as follows,
\begin{equation}
    \mathcal{L} = \frac{1}{\sum_{i=1}^{m} \sum_{j=1}^{c} \mathbf{y}_{i,j}} \sum_{i=1}^{m}\sum_{j=1}^{c} \textbf{FL}(\hat{\mathbf{y}}_{i,j}', \mathbf{y}_{i,j}),
\end{equation}
where 
$c$ is the number of action classes, $\mathbf{y}_{i,j} \in \{0, 1\}$ indicates whether the ground truth of the $i$-th human-object pair contains the $j$-th action class.

\textbf{Inference.} To make full use of the pre-trained object detector, we incorporate the object confidence scores into the final scores of each human–object pair as,

\begin{equation}
    \mathbf{s} = (s_h s_o)^{\lambda} \cdot \delta(\hat{\mathbf{y}}'),
\end{equation}
where hyperparameter $\lambda > 1$, $\delta(*)$ is the sigmoid function.

\begin{table*}[t]
\caption{Comparison with state-of-the-art methods on HICO-Det and V-COCO. \textbf{Bold} and \underline{underline} items indicate the best and second-best results, respectively. One-stage and two-stage methods are highlighted separately. }
\centering
\renewcommand{\arraystretch}{1.1}
\setlength{\tabcolsep}{6pt}  
\small  
\begin{tabular}{lccccccccc}
\hline
\multirow{3}{*}{Method} & \multirow{3}{*}{Backbone} & \multicolumn{3}{c}{Default (mAP \% $\uparrow$)} & \multicolumn{3}{c}{Known Object (mAP \% $\uparrow$)} & \multicolumn{2}{c}{V-COCO (\%)}  \\  \hline
 &  & Full & Rare & Non-Rare & Full & Rare & Non-Rare & $AP_{role}^{S_1}$ & $AP_{role}^{S_2}$  \\ 
\hline
\textbf{\textit{One-stage Methods:}}&  &  &  &  &  &  & & & \\ 
CATN\cite{dong2022category} \W{(ICCV’21)}  & R50 & 31.86 & 25.15 & 33.84 & 34.44 & 27.69 & 36.45 & 60.1 & - \\
GEN-VLKT \cite{liao2022gen} \W{(CVPR'2022)} & R50+ViT-B/32 & 33.75 & 29.25 & 35.10 & 36.78 & 32.75 & 37.99 & 62.4 & 64.5 \\
ERNet \cite{lim2023ernet} \W{(TIP'2023)} & EfficientNetV2-L & 34.25 & 28.70 & \textbf{36.33} & - & - & - & 61.6 & - \\
SG2HOI+ \cite{he2023toward} \W{(TIP'2023)} & R50+ViT-B/32 & 33.14 & 29.27 & 35.72 & 35.73 & 32.01 & 36.43 & \underline{63.6} & 65.2 \\
HODN \cite{fang2023hodn} \W{(TMM'2023)} & R50 & 33.14 & 28.54 & 34.52 & 35.86 & 31.18 & 37.26 & \textbf{67.0} & \textbf{69.1} \\
Multi-Step \cite{zhou2023learning} \W{(ACM MM'2023)} & R101 & 34.42 & 30.03 & 35.73 & 37.71 & 33.74 & \underline{38.89} & 61.3 & \underline{67.0} \\
HOICLIP \cite{ning2023hoiclip} \W{(CVPR'2023)} & R50+ViT-B/32 & 34.69 & 31.12 & 35.74 & 37.61 & 34.47 & 38.54 & 63.5 & 64.8 \\
CEFA \cite{zhang2024plug} \W{(ACM MM'2024)} & R50+ViT-B/32 & \underline{35.00} & \underline{32.30} & \underline{35.81} & \underline{38.23} & \underline{35.62} & \textbf{39.02} & 63.5 & - \\
DP-ADN \cite{gao2024dual} \W{(AAAI'2024)} & R50+ViT-B/32 & \textbf{35.91} & \textbf{35.82} & 35.44 & \textbf{38.99} & \textbf{39.61} & 38.80 & 62.6 & 64.8 \\
\hline
\textbf{\textit{Two-stage Methods:}}&  &  &  &  &  &  & & & \\ 
PViC\textsuperscript{$\dagger$}  \cite{zhang2023exploring} \W{(ICCV'2023)} & R50 & 34.69 & 32.14 & 35.45 & 38.14 & 35.38 & 38.97 & 59.7 & 65.4   \\ 
ADA-CM \cite{lei2023efficient} \W{(ICCV'2023)} & R50+ViT-L/14 & 38.40 & 37.52 & 38.66 & - & - & - & 58.6 & 64.0 \\ 
CLIP4HOI \cite{mao2023clip4hoi} \W{(NeurIPS'2023)} & R50+ViT-B/16 & 35.33 & 33.95 & 35.74 & - & - & - & - & 66.3 \\
CMMP \cite{lei2024exploring} \W{(ECCV'2024)}  & R50+ViT-L/14 & 38.14 & 37.75 & 38.25 & - & - & - & - & 64.0 \\
Pose-Aware \cite{wu2024exploring} \W{(CVPR'2024)}  & R50 & 35.86 & 32.48 & 36.86 & 39.48 & 36.10 & 40.49 & \textbf{61.1} & \underline{66.6} \\
EZ-HOI \cite{lei2024ez} \W{(NeurIPS'2024)}  & R50+ViT-L/14 & 38.61 & 37.70 & 38.90 & - & - & - & 60.5 & 66.2 \\
LAIN  \cite{kim2025locality} \W{(CVPR'2025)}  & R50+ViT-B/16 & 36.02 & 35.70 & 36.11 & - & - & - & - & 65.1 \\
HOLa \cite{lei2025hola} \W{(ICCV'2025)} & R50+ViT-L/14 & \underline{39.05} & \underline{38.66} & \underline{39.17} & - & - & - & 60.3 & 66.0 \\

\hdashline
\rowcolor{gray!30}
CRL-B (Ours) & R50+ViT-B/16 & 36.70 & 35.16 & 37.16 & \underline{40.17} & \underline{39.02} & \underline{40.51} & 60.2 & 65.9  \\
\rowcolor{gray!30}
CRL-L (Ours) & R50+ViT-L/14 & \textbf{39.99} & \textbf{40.67} & \textbf{39.78} & \textbf{43.35} & \textbf{44.43} & \textbf{43.02} & \underline{60.9} & \textbf{66.9}  \\
\hline
\multicolumn{10}{l}{$^\dagger$ The released code of PViC for V-COCO is no longer available, thus the results reproduced in \cite{wu2024exploring} are reported instead.} \\
\end{tabular}

\label{tab: hico}
\end{table*}

\begin{table}[t]
\centering
\caption{
Ablation study of our model. UA: unary association, TR: ternary relationship, GB: global block, RB: regional block. $\checkmark$ indicates that the module is used.
}
\renewcommand{\arraystretch}{1.2}
\begin{tabular}{ccccccc}
\hline
\multicolumn{2}{c}{\textbf{MRM}} & \multicolumn{2}{c}{\textbf{CPL}} & \multicolumn{3}{c}{\textbf{HICO-Det (Default)}} \\
\hline
UA & TR & GB & RB & Full & Rare & Non-Rare \\
\hline
- & - & - & - & 35.45 & 33.66 & 35.99 \\  
\hdashline
$\checkmark$ & - & - & - & 35.76 & \underline{34.10} & 36.25 \\ 
- & $\checkmark$ & - & - & 35.94 & 34.00 & 36.52 \\ 
$\checkmark$ & $\checkmark$ & - & - & 36.00& 34.07 & 36.58 \\ 
\hdashline
- & - & $\checkmark$ & - & 36.30 & 34.03 & 36.98 \\ 
- & - & - & $\checkmark$ & 35.85 & 33.25 & 36.63 \\ 
- & - & $\checkmark$ & $\checkmark$ & \underline{36.39} & 33.59 & \textbf{37.22} \\ 
\hdashline
$\checkmark$ & $\checkmark$ & $\checkmark$ & $\checkmark$ & \textbf{36.70} & \textbf{35.16} & \underline{37.16} \\ 
\hline
\end{tabular}
\label{tab:ablation_study}
\end{table}

\section{Experiments}

\subsection{Experiment Setup}
\textbf{Dataset.} We evaluate our model on two widely used benchmarks, HICO-DET \cite{chao2018learning} and V-COCO \cite{gupta2015visual}. HICO-DET consists of 37,633 training and 9,546 test images, covering 600 HOI categories derived from 80 object and 117 action classes, split into 138 rare and 462 non-rare categories. V-COCO is derived from COCO \cite{lin2014microsoft} and contains 10,326 images (5,400 for training, 4,964 for testing) with annotations for 80 object and 24 action categories.

{\textbf{Evaluation 
 Metrics.}} The mean average precision (mAP) is used to evaluate performance. A predicted HOI triplet is considered a true positive if it satisfies two conditions: 1) the Intersection over Union (IoU) between the predicted and ground-truth bounding boxes for both the human and the object exceeds 0.5; 2) the predicted action and object categories match the ground-truth labels.

\textbf{Zero-shot Setting.} Following prior works \cite{ning2023hoiclip,lei2024ez}, we conduct our zero-shot experiments under four distinct configurations: Rare First Unseen Combination (RF-UC), Non-rare First Unseen Combination (NF-UC), Unseen Verb (UV), and Unseen Object (UO). In the RF-UC setting, we select tail HOI categories as unseen categories, while in the NF-UC setting, we use head HOI categories as unseen categories. Under the UV and UO settings, some verb or object categories are not included in the training set, respectively.



{\textbf{Implementation Details.}} The pre-trained DETR model with a ResNet50 \cite{he2016deep} backbone is selected as our object detector. AdamW \cite{loshchilov2017decoupled} is used as the optimizer, with both the learning rate and weight decay being $10^{-4}$. We set $\lambda$ to 1 during training and 2.8 during inference. Unless otherwise specified, all models are trained for 15 epochs, with a learning rate drop by a factor of 5 at the $10^{th}$ epoch. The visual encoder is based on ViT-B/16 and ViT-L/14 CLIP, and during training, the parameters of CLIP remain frozen. In the MRM module, we use two blocks for both the binary and ternary decoders and set the weighting coefficient $\alpha$ to 1. \texttt{$[$PREFIX$]$} token is manually designed, with \texttt{$[$ACT$]$} token length set to 4. All experiments are conducted on 8 NVIDIA 4090 GPUs and the batch size is 16. The computational environment runs Ubuntu 22.04, with Python version 3.7, PyTorch version 1.10.0, torchvision version 0.11.0, and CUDA version 11.3.
\subsection{Comparison to the State-of-The-Art}

The experimental results on the HICO-Det and V-COCO datasets are presented in Table \ref{tab: hico}. We compared with both one-stage methods, such as CATN \cite{dong2022category}, ERNet \cite{lim2023ernet}, SGHOI+ \cite{he2023toward}, etc., and two-stage methods, like CMMP \cite{lei2024exploring}, EZ-HOI \cite{lei2024ez}, LAIN \cite{kim2025locality}, HOLa \cite{lei2025hola}. For the HICO-DET dataset, our proposed model demonstrates remarkable performance, outperforming the most recent work HOLa by a margin of \textbf{0.94} mAP for full categories. Notably, the performance improvement is most significant in the rare categories, where our model outperforms CMMP and EZ-HOI by margins of \textbf{2.92} mAP and \textbf{2.97} mAP, respectively. For the V-COCO dataset, our model achieves 60.9 and 66.9 role AP in S1 and S2, surpassing recent work HOLa by margins of \textbf{0.7} mAP and \textbf{0.9} mAP, respectively.



\begin{figure}[t]
\centering
\includegraphics[width=\columnwidth]{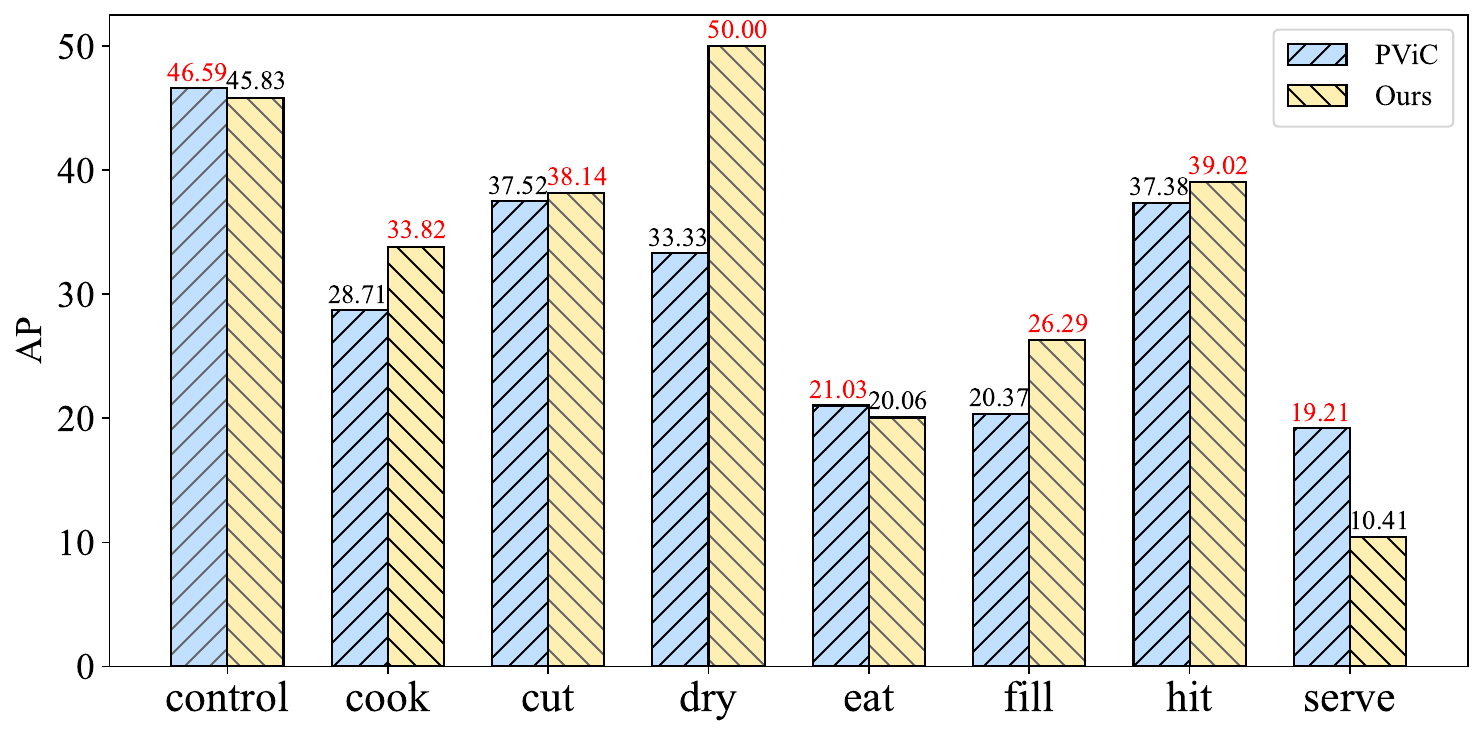}
\caption{Comparison of per-category accuracy between CRL-B and PViC \cite{zhang2023exploring} on HICO-Det-HTO.}

\label{fig:tool_ap}
\end{figure}

To further assess the impact of tool affordance-guided interaction modeling, we construct a benchmark subset from HICO-Det, named HICO-Det-HTO. This subset exclusively includes interactions where explicit tool usage is involved. The final subset includes 783 test images. A category-wise accuracy analysis for each action is conducted on this new dataset, allowing a comparison with the baseline method PViC \cite{zhang2023exploring}. As shown in Fig. \ref{fig:tool_ap}, it can be observed that our model outperforms PViC on several actions, including “cook”, “cut”, “dry”, “fill”, and “hit”.

\begin{table*}[t]
\centering
\caption{Ablative experiments for hyper-parameters in CPL. 
Manual: hand-crafted prompt. }
\footnotesize 
\setlength{\tabcolsep}{2.5pt} 
\begin{minipage}[b]{0.25\textwidth}  
\centering
\textbf{(a) Prompt length} \\
\begin{tabular}{cccc}
\hline
\textbf{[ACT]} & Full & Rare & Non-Rare \\ 
\hline
2  & 36.56 & 33.92 & \textbf{37.36} \\ 
\rowcolor{gray!30}
4  & \textbf{36.70} & \textbf{35.16} & 37.16 \\ 
6  & 36.05 & 33.88 & 36.69 \\ 
8  & 36.46 & 34.92 & 36.92 \\ 
10 & 36.57 & 34.62 & 37.15 \\ 
\hline
\end{tabular}
\end{minipage}%
\hspace{0.01\textwidth} 
\begin{minipage}[b]{0.41\textwidth}  
\centering
\textbf{(b) Prompt setting} \\
\begin{tabular}{ccccc}
\hline
\textbf{[PREFIX]} & \textbf{[ACT]} & Full & Rare & Non-Rare \\ 
\hline
\rowcolor{gray!30}
Manual & 4 & \textbf{36.70} & \textbf{35.16} & 37.16 \\ 
4      & 2 & 36.30 & 33.96 & 37.00 \\ 
4      & 4 & 36.28 & 34.64 & 36.76 \\ 
8      & 2 & 36.57 & 34.07 & \textbf{37.32} \\ 
8      & 4 & 36.54 & 34.41 & 37.17 \\ 
\hline
\end{tabular}
\end{minipage}%
\hspace{0.01\textwidth} 
\begin{minipage}[b]{0.23\textwidth}  
\centering
\textbf{(c) CPL weight} \\
\begin{tabular}{cccc}
\hline
$\beta$ & Full & Rare & Non-Rare \\ 
\hline
0.2  & 36.43 & 34.61 & 36.98 \\ 
\rowcolor{gray!30}
0.4  & \textbf{36.70} & 35.16 & \textbf{37.16} \\ 
0.6  & 36.49 & \textbf{35.26} & 36.85 \\ 
0.8  & 36.55 & 35.13 & 36.97 \\ 
1.0  & 36.54 & 35.13 & 36.96 \\ 
\hline

\end{tabular}

\end{minipage}

\label{tab:3abl}
\end{table*}

\subsection{Ablations Study}
To demonstrate the effectiveness of our framework, we conducted several ablation studies on the HICO-Det dataset. Noting that our baseline is built on the inferior Variant E3 of PViC \cite{zhang2023exploring}, which uses a ResNet-50 C5 backbone without an additional feature head. 
By replacing the decoder's original input feature with CLIP visual features $\mathbf{V}_e$,
this enhanced baseline achieves higher mAP scores compared to PViC, i.e., 35.45, 33.66 and 35.99 in full, rare and non-rare settings respectively.

{\textbf{Network Architecture Design.}}  As shown in Table \ref{tab:ablation_study}, 
the introduction of Multivariate Relationship Modeling (MRM) consistently delivers performance gains.
Concretely, it leads to \textbf{0.55} (full), \textbf{0.41} (rare) and \textbf{0.59} (non-rare) mAP improvements. In contrast, Contextualized Prompt Learning (CPL) achieves promising mAP increases in the full (\textbf{0.94}) and non-rare (\textbf{1.23}) settings. However, there is a slight decrease of \textbf{0.07} in mAP for the rare category. Fortunately, it can be effectively compensated by integrating the MRM module. When combined, they form a comprehensive model that delivers outstanding overall performance, notably achieving an impressive mAP boost of \textbf{1.50} in rare cases.


 


{\textbf{Multivariate Relationship Modeling.}} 
The modeled binary connection is inherently included in the baseline, which cannot be removed. Therefore, we focus on unary association (UA) and ternary relationship (TR) modeling.
As shown in the “MRM” column in Table \ref{tab:ablation_study}, both are effective in improving interaction recognition. For the Full, Rare and Non-rare metrics, associating the textual information in unary modeling can improve the mAP by \textbf{0.31}, \textbf{0.44} and \textbf{0.26}, respectively. Meanwhile, the sole use of ternary relationship modeling boost the mAP by \textbf{0.49}, \textbf{0.34}, and \textbf{0.53}. The best performance is achieved when both of them are combined.

{\textbf{Contextualized Prompt Learning.}} 
The proposed Contextual Decoder is crucial in CPL to refine textual features derived from learnable prompts. It comprises two blocks, namely the global block (GB) and the regional block (RB). 
As illustrated in the “CPL” column in Table \ref{tab:ablation_study}, both layers individually enhance HOI performance, and their combined use results in the most significant improvements.

Moreover, those predefined parameters in CPL also need to be carefully analyzed.
In Table \ref{tab:3abl} (a), we vary the length of learnable tokens \texttt{$[$ACT$]$} in our prompt template “\textit{A photo of a person \texttt{$[$ACT$]$} \{object\}}”. The best performance is achieved with 4 tokens, while using more than this could potentially increase the complexity of learning.
Instead of using the hand-crafted \texttt{$[$PREFIX$]$} tokens like “a photo of a”, we can use learnable ones, resulting in the format “\textit{\texttt{$[$PREFIX$]$} person \texttt{$[$ACT$]$} \{object\}}”. 
As shown in Table \ref{tab:3abl} (b), switching to learnable \texttt{$[$PREFIX$]$} tokens appears to be less effective. This suggests that having too much flexibility in prompts might be suboptimal, aligning with the findings presented in Table \ref{tab:3abl} (a). Finally, as given in Table \ref{tab:3abl} (c), the performance is not particularly sensitive to changes in the CPL weight (i.e., $\beta$).

\begin{figure}[t]
\centering
\includegraphics[width=\columnwidth]{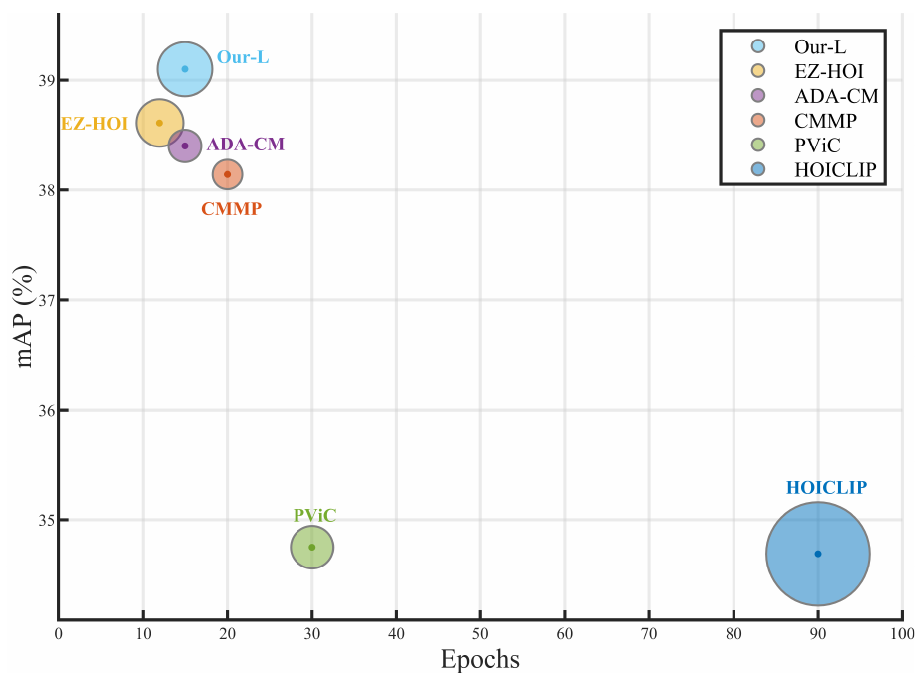}
\caption{Comparison of model performance with respect to learnable parameters and training epochs. }
\label{fig:model_complexity}
\end{figure}


\begin{figure*}[htbp]
\centering
\includegraphics[width=\textwidth]{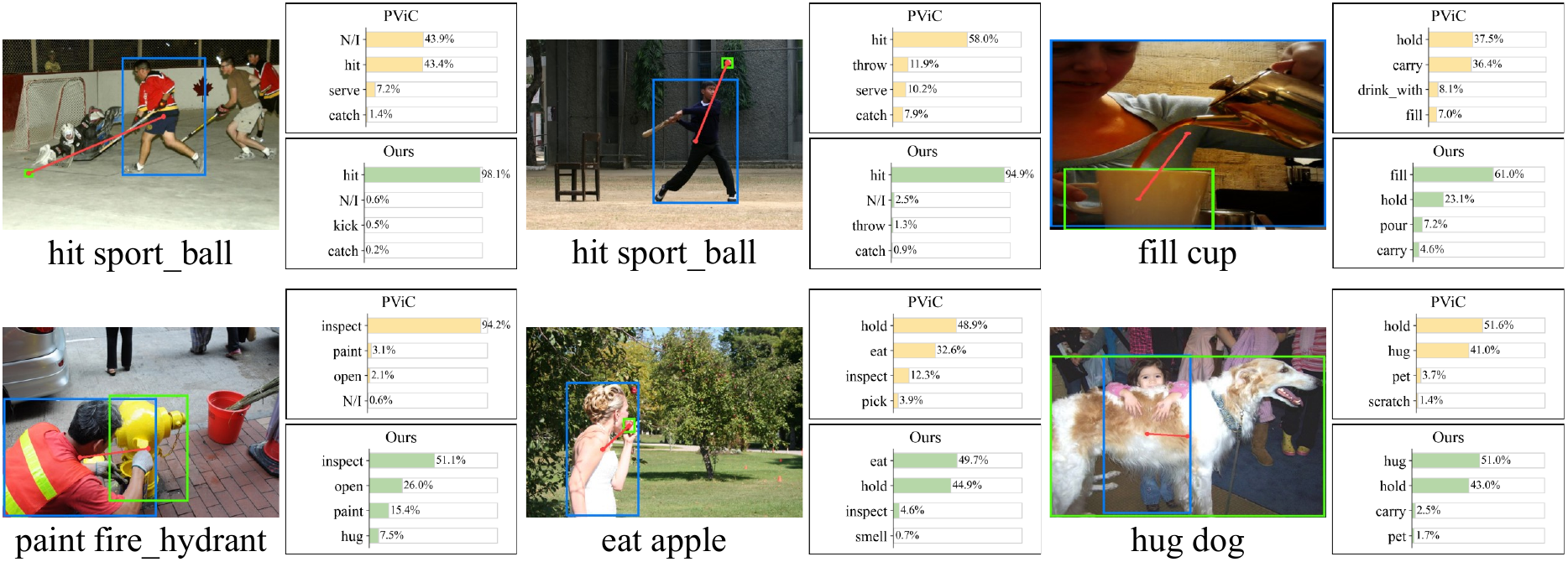}
\caption{Qualitative results on HICO-Det test set with fine-tuned DETR-R50 as the object detector. Bounding boxes of humans and
objects are drawn with blue and green boxes. The textual annotation below the figure represents the ground truth. N/I denotes no interaction.} 
\label{fig:vis_map}
\end{figure*}

\begin{figure}[t]
\centering
\includegraphics[width=\columnwidth]{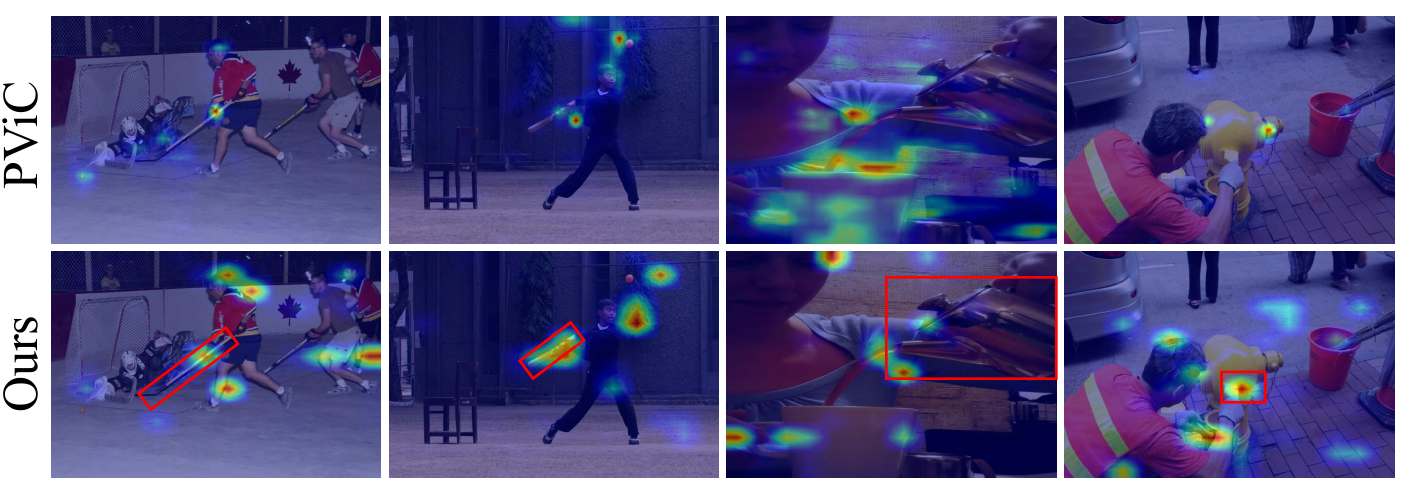}
\caption{The visualization of the cross-attention maps on a subset of images from Fig. \ref{fig:vis_map}. Bounding boxes of tools are drawn with red boxes. Best viewed in color. }
\label{fig:attn_map_tool}
\end{figure}

\subsection{Qualitative Results} As shown in Fig. \ref{fig:vis_map}, we present qualitative comparisons between our method and PViC \cite{zhang2023exploring} on the HICO-Det dataset. In the first row of the figure, we present three examples involving tool-related interactions. For actions related to “hit”, our model demonstrates more effective recognition of the “hit” interaction. Specifically, in the leftmost image of the first row, PViC tends to predict the interaction as “hold" or “carry", whereas guided by tool-related semantics, our model correctly identifies the interaction as “fill". It can be observed that our method performs robust in tool-related interaction cases and can consistently deliver more satisfying results. In the second row, we select examples of tool-irrelevant interactions for qualitative comparison. However, as illustrated by the rightmost example involving the action “paint", painting typically requires a tool such as a brush. Since the predefined object detector cannot recognize brushes (as “paintbrush" is not among the 80 COCO categories), we did not include this sample in the HICO-Det-HTO dataset. In this example, our model assigns a higher score proportion to the “paint" interaction, while PViC incorrectly classifies the interaction mostly as “inspect". In the two rightmost examples, PViC exhibits a tendency to classify the interactions as “hold”, while our approach successfully recognizes the more accurate interactions, namely “eat” and “hug”. 

Moreover, as shown in Fig.\ref{fig:attn_map_tool}, we visualize attention maps corresponding to some of the images in Fig.\ref{fig:vis_map}. Specifically, the first three examples illustrate the attention maps of the tool-related images in the first row of Fig.\ref{fig:vis_map}, where the attention is taken from the last layer of the ternary decoder. We observe that, compared to PViC, our model places greater focus on the tools involved in the interactions. The last image in Fig.\ref{fig:attn_map_tool} corresponds to the leftmost image in the second row of Fig.\ref{fig:vis_map}. As previously mentioned, the brush in this image cannot be detected by the object detector, and thus cannot be recognized as a tool. Consequently, the ternary token cannot be constructed, and no attention map is available from the ternary decoder. Instead, we visualize the attention map from the last layer of the pairwise decoder, and it shows that our model still attends to the brush held in the person’s hand.

During our experiments, we observed that existing datasets such as HICO-Det are not well-suited for annotating certain tool-related categories. For example, in some images depicting “a person hitting a baseball”, but the annotations are often limited to actions such as “person swinging a bat” or “person holding a bat”, which fail to capture the full semantics of the interaction. Moreover, the number of images involving measurable tool-related interactions is relatively small, making it difficult to effectively leverage tool-related information. In future work, we plan to expand the dataset by collecting more images, either from real-world sources or by generating synthetic images using generative models such as diffusion models.


\begin{table}[t]
    \caption{Zero-shot comparisons with SOTA methods on HICO-Det.}
    \centering
    \begin{tabular}{lcccc}
    \toprule
    Method & Type & Unseen & Seen & Full \\
    \midrule
    CMMP \cite{lei2024exploring} & RF-UC & 29.45 &  32.87 & 32.18 \\
    EZ-HOI \cite{ning2023hoiclip} & RF-UC & 29.02 & 34.15 & 33.13 \\
    LAIN \cite{kim2025locality} & RF-UC & \textbf{31.83} & 35.06 & \underline{34.41} \\
    HOLa \cite{lei2025hola} & RF-UC & 30.61 & \underline{35.08} & 34.19 \\
    Ours & RF-UC & \underline{31.73} & \textbf{36.61} & \textbf{35.63} \\
    \midrule
    CMMP \cite{lei2024exploring} & NF-UC & 32.09 &  29.71 & 30.18 \\
    EZ-HOI \cite{ning2023hoiclip} & NF-UC & 33.66 & 30.55 & 31.17 \\
    LAIN \cite{kim2025locality} & NF-UC & \textbf{36.41} & \underline{32.44} & \textbf{33.32} \\
    HOLa \cite{lei2025hola} & NF-UC & \underline{35.25} & 31.64 & \underline{32.36} \\
    Ours & NF-UC & 28.95 & \textbf{32.59} & 31.87 \\
    \midrule
    CMMP \cite{lei2024exploring} & UO & 33.76 & 31.15 & 31.59 \\
    EZ-HOI \cite{ning2023hoiclip} & UO & 33.28 & 32.06 & 32.27 \\
    LAIN \cite{kim2025locality} & UO & \textbf{37.88} & \underline{33.55} & \underline{34.27} \\
    HOLa \cite{lei2025hola} & UO & 36.45 & 33.02 & 33.59 \\
    Ours & UO & \underline{36.74} & \textbf{34.53} & \textbf{34.90} \\
    \midrule
    CMMP \cite{lei2024exploring} & UV & 26.23 &  32.75 & 31.84 \\
    EZ-HOI \cite{ning2023hoiclip} & UV & 25.10 & 33.49 & 32.32 \\
    LAIN \cite{kim2025locality} & UV & \textbf{28.96} & 33.80 & \underline{33.12} \\
    HOLa \cite{lei2025hola} & UV & \underline{27.91} & \underline{35.09} & \textbf{34.09} \\
    Ours & UV & 13.05 & \textbf{36.05} & 32.83 \\
    \bottomrule
    \end{tabular}
    
    \label{tab:zero-shot}
\end{table}

\subsection{Discussion}

\textbf{Model Complexity.} Considering that the introduction of the ternary decoder and contextual decoder introduces additional trainable parameters, we conduct a comprehensive comparison with several recent one-stage and two-stage HOI models. Specifically, we compare our method with EZ-HOI \cite{lei2024ez}, CMMP \cite{lei2024exploring}, and ADA-CM \cite{lei2023efficient}, all of which adopt ViT-L/14 as the backbone. We also include comparisons with our baseline PViC \cite{zhang2023exploring} and the one-stage method HOICLIP \cite{ning2023hoiclip}.

To provide a fair and holistic evaluation, we take into account not only model accuracy but also the number of learnable parameters and training epochs. As shown in Fig. \ref{fig:model_complexity}, all experiments are conducted on the HICO-DET test set. When using the CLIP ViT-L/14 variant, the visual encoder outputs 1024-dimensional features, which significantly increases the parameter size of the downstream attention modules. To mitigate this overhead, we adopt lightweight down-projection layers that reduce the feature dimensionality from 1024 to 512, effectively lowering the overall model complexity. As a result, our model contains 18.3M learnable parameters. While this is slightly higher than EZ-HOI (14.1M) and CMMP (5.4M), our method achieves better performance on the full set, reaching 39.10 mAP.

\textbf{Zero-shot Setting.} Table \ref{tab:zero-shot} compares the zero-shot HOI detection performance of our method with four SOTA approaches (CMMP, EZ-HOI, LAIN, HOLa) on HICO-Det across four scenarios (RF-UC, NF-UC, UO, UV), evaluating Unseen, Seen, and Full metrics. Since the experiment is conducted under a zero-shot setting, we freeze the last three classifiers and adopt the Verb Adapter from HOICLIP\cite{ning2023hoiclip} as the weights for these three classifiers. Our method exhibits notable strengths: it ranks first in the Seen metric across all scenarios (e.g., 36.61\% in RF-UC, 36.05\% in UV) and secures top positions in the Full metric for RF-UC (35.63\%) and UO (34.90\%), with competitive Unseen performance in these two scenarios (31.73\% for RF-UC, 36.74\% for UO). However, it has room for improvement: in NF-UC, its Unseen score (28.95\%) drags the Full metric to 31.87. In the unseen verb (UV) task, the low Unseen score (13.05\%) restricts the Full metric to only 32.83\%. The poor performance of unseen verbs stems from two key issues: tool-related ones like “stab” and “swing” are often misclassified as familiar actions such as “hit” or “serve” due to imprecise feature links, while abstract verbs like “inspect” and “install” lack consistent connections to known actions. Their vague boundaries and context-dependent use further hinder knowledge transfer. To fix this, future work should refine tool-action representations to capture subtle differences and integrate goal information for abstract verbs.

\section{Conclusion}
\label{sec:conclusion}
In this paper, we have proposed CRL, a two-stage HOI detection framework that addresses two key limitations of existing methods: the lack of tool-mediated modeling and the inability to integrate visual features into prompt learning, limiting their capacity to capture context-dependent relationships. Our approach has shown notable effectiveness in challenging HOI scenarios with complex tool-assisted interactions and ambiguous contextual cues, highlighting the benefits of multivariate relationship modeling and contextualized prompt learning. In future work, we plan to explore fine-grained affordance-guided modeling and broaden generalization to open-world HOI scenarios.

\bibliographystyle{IEEEtran}
\bibliography{main}

\end{document}